\documentclass[runningheads,a4paper]{llncs}

\usepackage{amssymb}
\usepackage{graphicx}
\usepackage{amsmath}
\usepackage{multirow}
\usepackage{url}
\urldef{\mailsa}\path|{gabriel.kronberger,michael.kommenda}@fh-hagenberg.at|
\newcommand{\keywords}[1]{\par\addvspace\baselineskip

\noindent\keywordname\enspace\ignorespaces#1}

\newcommand{\matern}{\text{Mat\'ern}}

\begin{document}

\mainmatter  
\title{Evolution of Covariance Functions for Gaussian Process Regression using Genetic Programming}

\titlerunning{Evolution of Covariance Functions for Gaussian Process Regression}

\author{Gabriel Kronberger \and  Michael Kommenda}
\authorrunning{Evolution of Covariance Functions for Gaussian Process Regression}
\institute{School of Informatics, Communications and Media,\\
University of Applied Sciences Upper Austria,\\
Softwarepark 11, 4232, Hagenberg\\
\mailsa}

\toctitle{Evolution of Covariance Functions for Gaussian Process Regression}
\tocauthor{Gabriel Kronberger, Michael Kommenda}
\maketitle

\begin{abstract}
In this contribution we describe an approach to evolve composite
covariance functions for Gaussian processes using genetic
programming. A critical aspect of Gaussian processes and similar
kernel-based models such as SVM is, that the covariance function
should be adapted to the modeled data.  Frequently, the squared
exponential covariance function is used as a default. However, this
can lead to a misspecified model, which does not fit the data well.

In the proposed approach we use a grammar for the composition of
covariance functions and genetic programming to search over the space
of sentences that can be derived from the grammar.

We tested the proposed approach on synthetic data from two-dimensional
test functions, and on the Mauna Loa $CO_2$ time series. The results
show, that our approach is feasible, finding covariance functions that
perform much better than a default covariance function. For the $CO_2$
data set a composite covariance function is found, that matches the
performance of a hand-tuned covariance function.
\end{abstract}

\section{Introduction}
% Short intro, mainly motivating the need to adapt covariance functions
% (should have a reference to the paper from Gelmans blog)
% And a sentence that Genetic programming can be potentially useful for this task

The composition of covariance functions is a non-trivial task and
has been described as a black art~\cite{Duvenaud2013}. On the one
hand, it is critical to tune the covariance function to the data set,
that should be modeled, because this is the primary option to
integrate prior knowledge into the learning
process~\cite{Rasmussen2006}; on the other hand a lot of experience
and knowledge about the modeled system is required to do this
correctly. Frequently, and especially for multi-dimensional data sets
it is far from obvious how the covariance function should be
structured.

% quick intro to Gaussian processes
In this work we discuss the composition of covariance functions for
Gaussian processes, that can be used for nonparametric machine
learning tasks e.g., for regression or
classification~\cite{Rasmussen2006}. In this context a Gaussian
process is used as a Bayesian prior over functions, relating the input
variables to the target variable. Gaussian process regression allows
modeling of non-linear functional dependencies through different
covariance functions, and produces posterior probability distribution
estimates for the target values instead of point estimates only.

% aims and non-aims of the paper
\subsection{Our Contribution}
% Short description of our contribution
The aim of this paper is to describe the idea of using a grammar for
covariance functions and genetic programming to search for a good
covariance function for a given data set. We also describe our
prototype implementation using grammar-guided tree-based GP, and
finally, present results as a proof-of concept. We have not yet
evaluated the difficulty of this problem for genetic programming, and
in particular, if GP suited well for this kind of problem. The results
of our experiments indicate that the idea is feasible, producing good
covariance functions for low-dimensional data sets.

\subsection{Previous Work}
In a very recent contribution, the problem of structure identification
for covariance functions has been approached using a grammar, or
rather a set of rewriting rules, as a basis for searching over
composite covariance functions for Gaussian
processes~\cite{Duvenaud2013}. This approach is actually very similar
to our work; the main difference is, that in our work we use genetic
programming to search over the set of possible structures, while
Duvenaud et al. enumerate over composite functions, starting with
standard functions. 
%The differences of our approach compared to this
% work are discussed in more detail in the final section of this paper.

Another recent contribution discusses more flexible families of
covariance functions, instead of composing covariance functions from
simple terms~\cite{Wilson2013}. Also related is earlier work that
describes additive Gaussian processes~\cite{Duvenaud2011}, which are
equivalent to a weighted additive composition of base kernels, but can
be calculated efficiently.

Genetic programming has been used previously to evolve kernel
functions for SVMs with mixed
results~\cite{Gagne2006},~\cite{Howley2006}. The latest contribution
found that genetic programming was able to \emph{``rediscover multiple
  standard kernels, but no significant improvements over standard
  kernels were obtained''}~\cite{Koch2012}.  These results can,
however, not be transfered directly to Gaussian processes because of
several major differences between Gaussian processes and SVMs. In
particular, in the case of Gaussian processes hyper-parameters are
optimized using a ML-II approach, in contrast to SVMs, where
hyper-parameter values are usually tuned using cross-validation and
grid-search. Additionally, in contrast to all other previous work,
simple embeddings of covariance functions by masking dimensions are
supported.

\section{Gaussian Processes}
% more details on Gaussian Processes
A Gaussian process is a non-parametric model that produces predictions
solely from the specified mean and covariance functions and the
available training data~\cite{Rasmussen2006}. The inference of
function values $\vec{f}^*$ for observed input values $X^*$ based on
observations of $\vec{y}$ and $X$ involves the calculation of the
covariance matrices $K(X,X)$ and $K(X,X^*)$ and inference from the
multi-dimensional Gaussian shown in Equation \ref{eqn:gp_post}.
\begin{equation}
\left[
\begin{array}{c}
\vec{y} \\
\vec{f}^*
\end{array}
\right] \sim N \left(
\left[
\begin{array}{c}
m(X)\\
m(X^*)
\end{array}
\right]
,
\left[
\begin{array}{cc}
K(X,X) + \sigma^2 I & K(X, X^*) \\
K(X^*, X) & K(X^*, X^*)
\end{array}
\right]
\right)
\label{eqn:gp_post}
\end{equation}

The term $\sigma^2 I$ is necessary to account for Gaussian distributed
noise with variance $\sigma^2$.  From this definition it follows that
the posterior for $\vec{f}^*$ is again a multi-dimensional Gaussian.
For model selection and hyper-parameter learning the marginal
likelihood $p(\vec{y}|X)$ must be calculated. The model is a
multi-dimensional Gaussian so an analytical form of the likelihood can
be derived. 

Calculation of the marginal likelihood requires a matrix inversion
and, thus, has asymptotic complexity $O(n^3)$. Usually, the covariance
function $K(x,x')$ has hyper-parameters that must optimized. This is
often accomplished in a simple ML-II fashion, optimizing the
hyper-parameters w.r.t. the likelihood using a quasi-Newton method
(e.g., BFGS). Since the gradients of the marginal likelihood for the
hyper-parameters can be determined with an additional computational
complexity of $O(n^2)$ for each hyper-parameter, it is feasible to use
gradient-based methods. The drawback is that the likelihood is
typically multi-modal, and especially for covariances with many
hyper-parameters (e.g., ARD) the optimizer can converge to a local
optimum. Thus, it is typically suggested to execute several random
restarts. A better solution would be to include priors on the
hyper-parameters and optimizing w.r.t. posterior
distribution (MAP). However, this can only be accomplished using a MCMC
approach which is computationally expensive.

Frequently used covariance functions for Gaussian processes include
the linear, polynomial, squared exponential~(SE), rational
quadratic~(RQ) and the Mat\'ern function. Covariance functions can be
combined to more complex covariance functions, for instance as products
or sums of different covariance functions~\cite{Rasmussen2006}.

% Figure with different realizations of GPs

\section{Genetic Programming}
% Short intro to genetic programming
Genetic programming generally refers to the automatic creation of
computer programs using genetic algorithms~\cite{Koza1992}. The basic
principle is to evolve variable-length structures, frequently symbolic
expression trees, which represent potential solutions to the
problem. One of the most prominent applications of genetic programming
is symbolic regression, the synthesis of regression models without
a predetermined structure. Genetic programming makes it possible to
optimize the structure of solutions in combination with their
parameters. Thus, it should also be possible to synthesize composite
covariance functions with genetic programming.
In the following, we use a grammar-guided genetic programming system to
make sure that only valid covariance functions are produced. A good
survey of grammar-guided genetic programming is given in~\cite{McKay2010}.

\section{Grammar for Covariance Functions}
The grammar for covariance functions has been derived from the rules
for the composition of kernels as e.g., discussed
in~\cite{Rasmussen2006}. It should be noted that the grammar shown
below is not complete, meaning that several constructions that would
lead to a valid covariance function are not possible\footnote{One
  example is vertical scaling of covariance functions: $K'(x,x') = a(x)
  K(x,x') a(x')$}. The following represents the grammar
$G(\text{Cov})$ for covariance functions in EBNF notation\footnote{The
  grammar is largely based on the capabilities of the GPML package by
  Rasmussen and Nickisch, \url{http://gaussianprocess.org/gpml/code}.}: 
{\footnotesize
\begin{verbatim}
Cov         -> "Prod" "(" Cov { Cov } ")" | "Sum"  "(" Cov { Cov } ")" | 
               "Scale" Cov | "Mask" BitVector Cov | TerminalCov  . 
TerminalCov -> "SE" | "RQ" | "Matern1" | "Matern3" | "Matern5" |
               "Periodic" | "Linear" | "Constant" | "Noise" . 
BitVector   -> "[" {"0" | "1" } "]" . 
\end{verbatim}
}

The functions \verb+Prod+ and \verb+Sum+ produce the product and sum
of multiple covariance functions, which can again be composite
covariance functions.  The scale operator can be used to add a scaling
factor to any covariance function.  The \verb+Mask+ operator selects a
potentially empty subset of input variables from all possible input
variables.  The non-terminal symbol \verb+BitVector+ can be derived to
a list of zeros and ones. The bit vector is used to mask selected
dimensions in the data set, effectively reducing the
dimensionality. The length of the bit mask has to match the total number
of dimensions; this is checked when the resulting covariance function
is evaluated.

Finally, the non-terminal symbol \verb+TerminalCov+ can be derived to
a range of default covariance functions. Currently, we only included
isometric covariance functions, but other covariance functions can be
added to the grammar easily. The grammar does not include the
hyper-parameters, because they are not optimized by genetic
programming. Instead, hyper-parameters are optimized for each
potential solution, using a gradient-descent technique. 
%An exemplary
%sentence that can be derived from above grammar is:
%\begin{verbatim}
%Sum ( SEiso Sum ( Mask [01001] RQiso  Mask [10110] SEiso )) .
%\end{verbatim}

\section{Experiments}
For the experiments we implemented Gaussian processes, a set of
commonly used covariance functions, and the grammar for covariance
functions in HeuristicLab\footnote{HeuristicLab version 3.3.8 is
  available from
  \url{http://dev.heuristiclab.com/}}~\cite{Wagner2009} which already
provides an implementation of grammar-guided tree-based genetic
programming. 

The aim of the experiments presented in this contribution is mainly to
test the feasibility of the idea. Two different types of data sets are
used for the experiments, and the forecasts of the synthesized
covariance functions are compared to a set of default covariance
functions and also to hand-tuned covariance functions. The first data
set is the univariate Mauna Loa atmospheric $CO_2$ time series. This
data set has been chosen, because a hand-tuned covariance function for
this data set is presented in~\cite{Rasmussen2006}. For the second
experiment we created several synthetic data sets sampled randomly
from two-dimensional Gaussian process priors shown in
Equation~\ref{eqn:benchmark}. The data generated from these functions
are difficult to model with a single isometric covariance
function. Multiple covariance functions have to be combined and the
correct dimension masking vectors have to be identified. Each data set
contains 882 samples of the function on a regular two-dimensional
grid.

\begin{equation}
\begin{split}
\text{SE+RQ}(\vec{x},\vec{x'}) & = \text{SE}(x_0,x'_0) + \text{RQ}(x_1,x'_1)\\ 
\text{SE+Mat\'ern}(\vec{x},\vec{x'}) & = \text{SE}(x_0,x'_0) + \text{Mat\'ern1}(x_1,x'_1)\\ 
\text{SE+Periodic}(\vec{x},\vec{x'}) & = \text{SE}(x_0,x'_0) + \text{RQ}(x_1,x'_1)
\end{split}
\label{eqn:benchmark}
\end{equation}

%%\begin{tikzpicture}
%%\begin{axis}
%%\addplot3[surf] file {plotdata/se+rq.dat};
%%\end{axis}
%%\end{tikzpicture}
%%
% XXX TODO make pgfplots of the data sets

\subsection{Genetic Programming Parameter Settings}
Training of Gaussian processes is computationally expensive, and
because it is necessary to optimize the hyper-parameters for each
evaluated covariance function the run time of the genetic programming
algorithm grows quickly. Therefore, we used very restrictive
parameter settings, in particular a small population size of only 50
individuals. All other parameter settings are shown in
Table~\ref{tab:gp-settings}.
% Table for GP Settings
\begin{table}
\caption{\label{tab:gp-settings}Genetic programming parameter settings for all experiments.}
\centering
\begin{tabular}{l|l}
\textbf{Parameter} & \textbf{Value} \\
\hline
Population size & 50 \\
Max. length / height & 25 / 7 \\
Initialization & PTC2 \\
Parent selection & gender-specific (proportional + random ) \\
Mutation rate & 15\% \\
ML-II iterations & 50 \\
Offspring selection~\cite{Affenzeller:GAGP} & strict (success ratio = 1, comparison factor = 1) \\
Max. selection pressure & 100 \\
Max. generations & 20 \\
\end{tabular}
\end{table}

\subsection{Results on Mauna Loa $CO_2$ Data Set}
\begin{figure}
\includegraphics[width=\textwidth]{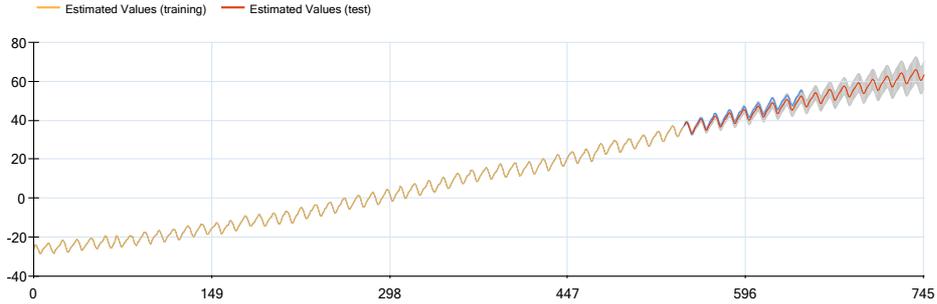}
\caption{\label{fig:co_model1}The output and forecast for the Mauna
  Loa $CO_2$ time series of a Gaussian process using the first evolved
  covariance function ($K1$) shown in Equation \ref{eqn:model}.}
\end{figure}

The results for the $CO_2$ time series are positive. The algorithm was
able to consistently find covariance functions that fit well in the
training period (1958~--~2004), accurate forecasts over the testing
period (2004~--~2012). The structures of two exemplary solutions are
shown in Equation~\ref{eqn:model}. The first solution ($K1$) is
actually very similar to the hand-tuned covariance solution proposed
in~\cite{Rasmussen2006}. The second covariance function is more
complex and has only a slightly better likelihood. Unfortunately,
genetic programming often leads to overly complex solutions which is a
critical drawback of our approach.  Both solutions have been found
after only 800 evaluated solution candidates and achieve a negative
log-likelihood of 129.8 and 116, respectively. The correlation
coefficients for the forecasts in the test partition are above
0.99. Figure~\ref{fig:co_model1} shows the output of the first model.

\begin{equation}
\begin{split}
% 129.8, R² (test): 0.993
K1(x,x') =\: & \text{SE}(x,x') + \text{Periodic}(x,x') + \matern1(x,x') + \\
            & \text{SE}(x,x') + \matern5(x,x') + \text{Const} \\
% 116, R² 0.994
K2(x,x') =\: & \matern3(x,x') * \text{Perioric}(x,x') * \text{RQ}(x,x')\: * \\
          & (\matern1(x,x') + \matern3(x,x') + \matern5(x,x')\: + \\ 
          & \: \text{Perioric}(x,x') + \text{Linear}(x,x'))\:* \\
          & (\matern1(x,x') + \matern3(x,x') + \text{RQ}(x,x'))
\end{split}
\label{eqn:model}
\end{equation}

\subsection{Results on Synthetic Data Sets}
The results for the synthetic two-dimensional data set are shown in
Table~\ref{tab:results-2d}. In this experiment we trained multiple
Gaussian process models using several frequently used covariance
functions. We trained many models using random restarts for each data
set and covariance function, and report the best negative
log-likelihood for each pair. As expected, the models with the
isometric covariance functions do not fit well. In contrast, the
composite covariance functions produced by genetic programming fit
much better. For comparison, we also report the negative
log-likelihood, that can be achieved with the optimal covariance
function for each data set. In these experiments, the exact structure
of the covariance could not be rediscovered, thus, the evolved
functions are worse than the optimal solution.
\begin{table}
\caption{\label{tab:results-2d}Best negative log-likelihood achieved for the three synthetic two-dimensional test functions, with default covariance functions and with evolved composite covariance functions.}
\centering
\begin{tabular}{l | l | r | r | r }
 \multicolumn{2}{c|}{}  & \multicolumn{3}{c}{\textbf{Problem instance}} \\
 \multicolumn{2}{c|}{}   & \textbf{SE+RQ} & \textbf{SE+Mat\'ern} & \textbf{SE+Periodic} \\
\hline
\multirow{6}{*}{\rotatebox{90}{\textbf{Covariance}}} & SE & -204 & -492 & 440 \\
     & RQ & -272 & -492 & 103 \\
     & Periodic & -221 & -492 & 479 \\
     & Mat\'ern & -27 & 31 & 304 \\
     & Evolved & -803 & -760 & -640 \\
     & Optimal & -2180 & -2187 & -2131 \\
\end{tabular}
\end{table}

\section{Summary and Discussion}
% Fully explain the differences compared to Duvenaud
In this contribution we described an approach for the synthesis of
composite covariance function for Gaussian processes using
grammar-guided genetic programming. In the proposed approach a set of
commonly used covariance functions is used to compose more complex
covariance functions, using sums or products of several covariance
functions. The set of valid covariance functions is defined via a
grammar and genetic programming is used to search the space of
possible derivations from this grammar. The hyper-parameters of
covariance functions are not subject to the evolutionary search, but
are optimized w.r.t. the likelihood using a standard gradient-descent
optimizer (i.e., LBFGS).

The proposed approach was tested on two types of low-dimensional
problems as a proof of concept. We found, that for the univariate
Mauna Loa $CO_2$ time series it is possible to consistently find good
covariance functions with genetic programming. The identified
solutions perform as well as a hand-tuned covariance function for this
problem. The results for our two-dimensional synthetic functions show
that it is possible to find composite covariance functions, which
perform much better than default covariance functions on these data
sets.

In contrast to previous work by the genetic programming
community~\cite{Koch2012}, which focused mainly on kernel synthesis
for SVMs, this contribution discusses kernel synthesis for Gaussian
processes, which are non-parametric fully Bayesian models. For
Gaussian process models the hyper-parameters can be optimized with a
standard gradient-descent approach, and it is not strictly necessary
to execute cross-validation~\cite{Rasmussen2006}. Previous work either used
grid-search and cross-validation to tune hyper-parameters, which is
very computationally expensive, or did not consider hyper-parameter
optimization at all. Additionally, we are using a grammar to compose
covariance functions from simple covariance functions instead of
evolving the full function.

In the statistics community, a very recent contribution has also
discussed the usage of grammars for the composition of covariance
functions~\cite{Duvenaud2013}. The main difference to this work is
that here genetic programming is used to search over the derivations
of the grammar. Another relevant difference is that the grammar in
this contribution also supports simple embeddings through the masking
function.  It should be noted, that we have not yet analyzed if
genetic programming is well suited for this task, and in particular we
did not compare the approach to simple enumeration or random search.

One question that remains for future work is whether composed
covariance functions also work well for data sets with more
variables. We have observed that simple covariance functions often
work very well, and tuned covariance functions
do not have a strong beneficial effect for these data sets.

Another interesting topic for future research is to look at
alternative ways for searching over the space of covariance functions
defined by a grammar. Recently, an interesting approach has been
described that uses variational methods for Bayesian learning of
probabilistic context free grammars for this
task~\cite{Hasegawa2009}. This idea could be especially useful for
Bayesian models such as Gaussian processes.

\subsubsection*{Acknowledgments}
The authors would like to thank Jeffrey Emanuel for the initial idea
leading to this contribution. This work has been supported by the Austrian
Research Promotion Agency (FFG), on behalf of the Austrian Federal
Ministry of Economy, Family and Youth (BMWFJ), within the program
``Josef Ressel-Centers''.

%%\bibliographystyle{splncs03}
%%\bibliography{kronberger}

\end{document}